\pdfoutput=1

\documentclass[11pt]{article}

\usepackage[]{ACL2023}

\usepackage{times}
\usepackage{latexsym}

\usepackage[T1]{fontenc}

\usepackage[utf8]{inputenc}

\usepackage{microtype}

\usepackage{inconsolata}

\usepackage{array}
\usepackage{pifont}
\usepackage{tabularx}
\usepackage{adjustbox}
\usepackage{multirow}
\usepackage{enumitem}
\usepackage{xspace}
\usepackage{tcolorbox}
\usepackage{booktabs,subcaption,amsfonts,dcolumn}
\usepackage{url}
\usepackage{amsmath,amsthm,amsfonts,amssymb,bm,stmaryrd}
\usepackage{color,xcolor,colortbl}
\usepackage{CJKutf8}
\usepackage{makecell}
\usepackage{booktabs}
\usepackage{graphicx}
\usepackage{listings}

\definecolor{mytype}{RGB}{91,155,213}
\definecolor{myentity}{RGB}{237,125,49}
\definecolor{relation}{RGB}{204,51,0}

\title{How to Unleash the Power of Large Language Models for \\
Few-shot Relation Extraction?}

\author{
Xin Xu, 
Yuqi Zhu,
Xiaohan Wang,
Ningyu Zhang\textsuperscript{\thanks{\quad Corresponding author.}} \\
Zhejiang University \& AZFT Joint Lab for Knowledge Engine\\
 \{xxucs, wangxh07, zhangningyu\}@zju.edu.cn
}

\begin{document}
\maketitle
\begin{abstract}
Scaling language models have revolutionized widespread NLP tasks, yet little comprehensively explored few-shot relation extraction with large language models. In this paper, we investigate principal methodologies, in-context learning and data generation, for few-shot relation extraction via GPT-3.5 through exhaustive experiments. To enhance few-shot performance, we further propose task-related instructions and schema-constrained data generation. We observe that in-context learning can achieve performance on par with previous prompt learning approaches, and data generation with the large language model can boost previous solutions to obtain new state-of-the-art few-shot results on four widely-studied relation extraction datasets. We hope our work can inspire future research for the capabilities of large language models in few-shot relation extraction\footnote{Code is available in \url{https://github.com/zjunlp/DeepKE/tree/main/example/llm}.}.
 
\end{abstract}

\section{Introduction}
 
Few-shot Relation Extraction (RE) appeals to many researchers in Natural Language Processing (NLP) due to the capability to extract textual information where only a few support examples are given \cite{han-etal-2018-fewrel,DBLP:conf/acl/YangZNZP20,DBLP:conf/emnlp/Han0L21,brody-etal-2021-towards,DBLP:journals/corr/abs-2303-08559}.
Most previous works focus on fine-tuning \cite{DBLP:conf/acl/SoaresFLK19,ye-etal-2022-generative} or prompt-tuning \cite{knowprompt,PTR} with relatively small language models, e.g., RoBERTa \cite{DBLP:journals/corr/abs-1907-11692}.
Recently, with the scaling of model size and corpus size, large language models (LLMs) such as ChatGPT \cite{ChatGPT-OpenAI} and GPT-4 \cite{GPT-4} have demonstrated powerful abilities by demonstrating only a few instances, a.k.a In-Context Learning \cite{DBLP:journals/corr/abs-2301-00234}. 
Although LLMs have achieved remarkable results in many NLP tasks, their potential in few-shot relation extraction has not been fully explored yet.

In this paper, we take GPT-3.5 \cite{gpt3.5} as an exemplary LLM to investigate how to maximize the utilization of LLMs for the few-shot relation extraction task with in-context learning and data generation.
Different from text classification, the relation extraction task contains rich pre-defined schemas (e.g., entity and relation type constraints) and a relatively large and complex classification space with noisy data.   
We further design two simple-yet-effective strategies to unleash the power of large language models better: \textbf{task-related instructions} and \textbf{schema-constrained data generation}.
We conduct exhaustive experiments on four well-known relation extraction datasets.
Empirical results indicate that LLMs can potentially be advantageous to few-shot relation extraction and boost previous prompt learning performance.
 
\begin{figure*}
     \includegraphics[width =1\linewidth]{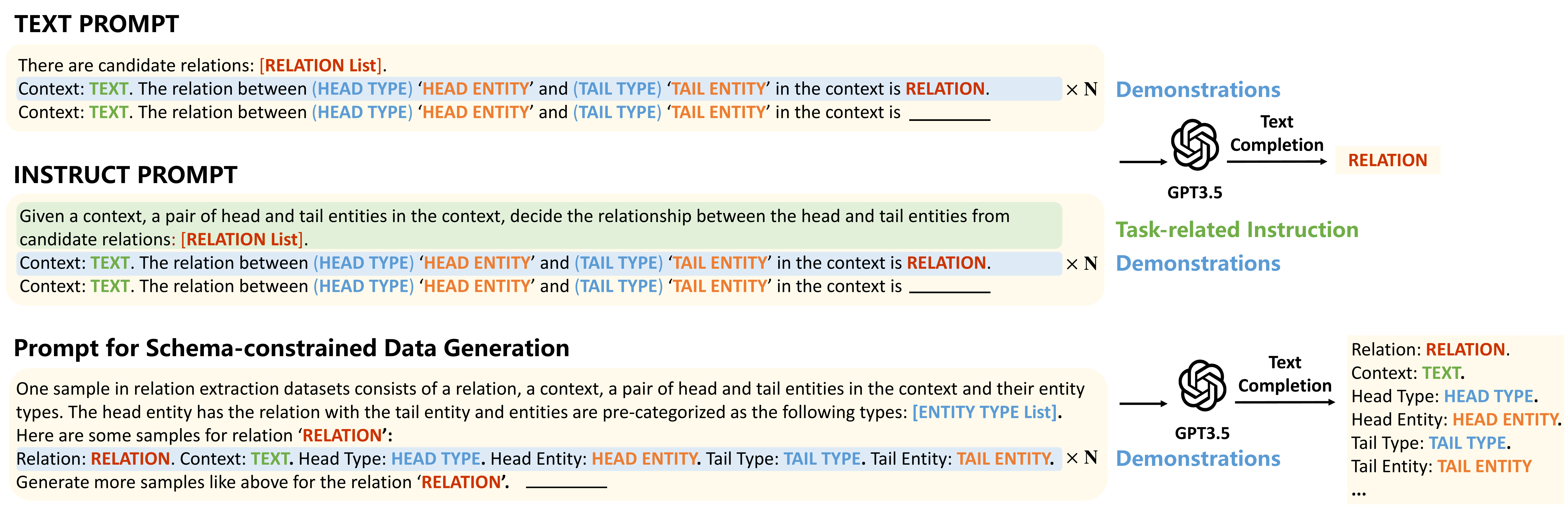}
    \caption{Strategies to unleash the power of LLMs for few-shot relation. \textcolor{mytype}{\textsc{head type}} and \textcolor{mytype}{\textsc{tail type}} are schemas. \textcolor{myentity}{\textsc{head entity}} and \textcolor{myentity}{\textsc{tail entity}} are entity mentions. \textcolor{relation}{\textsc{relation}} refers the verbalized relation label words. }
    \label{fig:intro}
\end{figure*}

\section{Background}
\subsection{Few-shot Relation Extraction}

The relation extraction task aims to extract the relationship between head and tail entities within a plain context.
Specifically, one instance for the relation extraction task consists of a context $\boldsymbol{x}=\{x_1, x_2, ..., h, ..., t, ..., x_{|\boldsymbol{x}|}\}$, head and tail entity mentions $\boldsymbol{h}$ and $\boldsymbol{t}$, entity types $\boldsymbol{t}_h$ and $\boldsymbol{t}_t$, and the relation $\boldsymbol{y} \in \mathcal{Y}$ between $\boldsymbol{h}$ and $\boldsymbol{t}$, where $\mathcal{Y}$ is the set of candidate relations. RE systems will predict $\boldsymbol{y}$ given $\boldsymbol{x}, \boldsymbol{h}, \boldsymbol{t}, \boldsymbol{t}_h$ and $ \boldsymbol{t}_t$.
For few-shot relation extraction, fine-tuning pre-trained language models (PLMs) is a direct solution \cite{opennre,luke,spanbert,DBLP:conf/acl/LyuC21,typmarker}. 
To alleviate the gap between pre-training objectives and downstream applications, prompt tuning has recently been applied to relation extraction, especially for low-resource scenarios \cite{knowprompt,PTR,genpt}. 
Most of those approaches utilize relatively small language models (RoBERTa \cite{DBLP:journals/corr/abs-1907-11692}, GPT2 \cite{radford2019language}), demonstrating empirical success regarding few-shot relation extraction performance. 
To date, large language models have demonstrated powerful abilities by prompting a few instances without tuning \cite{DBLP:journals/corr/abs-2212-10450}; however, the power of LLMs for few-shot relation extraction is little known.

\subsection{Large Language Models}
Large language models, trained with exceedingly large corpora and often with a great number of parameters ($\geq$10B), have achieved excellent performance in numerous downstream NLP tasks \cite{DBLP:journals/corr/abs-2211-09085, DBLP:journals/corr/abs-2205-01068, DBLP:journals/corr/abs-2210-02414,DBLP:journals/corr/abs-2204-02311, instructGPT}.
Compared to relatively small language models (SLMs), LLMs are usually not open-source and can not be fine-tuned, which is challenging for downstream task adaptation.
Therefore, in-context learning \cite{gpt3} is proposed to utilize prompts with a few demonstrations for few-shot learning.
Previous studies \cite{DBLP:conf/emnlp/YooPKLP21,DBLP:conf/emnlp/WangLXZZ21} have investigated using LLMs for text classification and generation. 
In this work, we take the first step to study few-shot RE with large language models, which brings new challenges and insights.

\section{LLMs for Few-shot Relation Extraction}
In this section, we introduce two strategies to utilize LLMs for relation extraction: 1) in-context learning (\S \ref{sec:in-context}); 2) data generation (\S \ref{sec:DA}) with LLMs, as shown in Figure \ref{fig:intro}.

\subsection{In-Context Learning with LLMs}
\label{sec:in-context}

\begin{table*}
  \centering
  \scalebox{0.9}{
    \begin{tabular}{clcccccccc}
    \toprule
    \multicolumn{2}{c}{\multirow{2}[2]{*}{Method}} & \multicolumn{2}{c}{TACRED} & \multicolumn{2}{c}{TACREV} & \multicolumn{2}{c}{RE-TACRED} & \multicolumn{2}{c}{SciERC} \\
    \multicolumn{2}{c}{} & K=8 & K=16 & K=8 & K=16 & K=8 & K=16 & K=8 & K=16 \\
    \midrule
    \multirow{6}[2]{*}{\rotatebox{90}{Baselines}} & SpanBERT \cite{spanbert} & 8.4 & 17.5 & 5.2  & 5.7  & 14.2  & 29.3  & 29.0 & 38.7 \\
      & LUKE \cite{luke} &  9.5 & 21.5  & 9.8  & 22.0  & 14.1  & 37.5  & 33.2  & 48.9 \\
      & GDPNet \cite{gdpnet} & 11.8  & 22.5  & 8.3  & 20.8  & 18.8  & 48.0  & 33.5  & 42.3 \\
      & TANL \cite{tanl} & 18.1  & 27.6  &  18.6 & 28.8  & 26.7 & 50.4  & 32.4  & 38.7 \\
      & TYP Marker \cite{typmarker} & 26.5 & 29.9 &  26.7  & 29.5  & 44.8  & 54.1 & 50.4  & 59.0 \\
      & KnowPrompt \cite{knowprompt} & 29.4 & 32.1 &  29.8  & 34.1  & 56.1  & 61.4  & 50.2  & 57.1  \\
    \midrule
    \multirow{4}[2]{*}{\rotatebox{90}{GPT3}} & In-context Learning\dag & \multicolumn{2}{c}{31.9}  & \multicolumn{2}{c}{32.4}  & \multicolumn{2}{c}{49.9}  & \multicolumn{2}{c}{46.6}  \\
    & In-context Learning\dag  (w/ Instruction) & \multicolumn{2}{c}{31.0}  &\multicolumn{2}{c}{31.9}  & \multicolumn{2}{c}{51.8}  & \multicolumn{2}{c}{48.8}  \\
    \cmidrule{2-10}
    &  Data Generation (TYP Marker) & 35.8  & 36.6  & 36.7  & 36.5  & 58.4  & 60.6  & 63.2  & 64.3  \\
    & Data Generation (KnowPrompt) & 37.9  & 37.4  & 42.6  & 41.0  & 62.7  & 66.2  & 58.6  & 67.8  \\
      
    \bottomrule
    \end{tabular}}
    \caption{Micro F1 (\%) of few-shot performance. $\dag$ refers to the performance with one-shot demonstrations.}
  \label{tab:main}
\end{table*}

\begin{table}[]
  \centering
  \scalebox{0.65}{
    \begin{tabular}{ccccc}
    \toprule
    Prompts & TACRED & TACREV & RE-TACRED & SciERC \\
    \midrule
    TEXT & 31.9 & 32.4  & 49.9  & 46.6  \\
    TEXT + Schema & 36.9 & 37.7 & 54.3  & 45.9  \\
    INSTRUCT & 31.0  & 31.9  & 51.8  & 48.8  \\
    INSTRUCT + Schema & 38.3 & 36.7 & 58.5  & 50.2  \\
    \bottomrule
    \end{tabular}}
    \caption{Micro F1 (\%) of performance on different prompt: \textsc{text prompt} and \textsc{instruct prompt}.}
  \label{tab:prompt}
\end{table}

The first strategy applies in-context learning (ICL) by providing LLMs with demonstrations in the prompt to elicit comprehension of the relation extraction task from LLMs.
To this end, specific and compelling prompts for RE with demonstrations are manually constructed and designed to instruct LLMs to understand the relation extraction task and how to execute relation extraction.
Considering aspects and characteristics of the relation extraction task, including task definition, candidate relation (label) words, entity types (schemas) and so on, we design prompts of different articulation and complexity to investigate how prompts help LLMs release the power of few-shot RE.
First, \textbf{\textsc{text prompt}} only contains essential elements for RE, including relation categories, contexts, and corresponding head and tail entities.
Inspired by the fantastic performance of InstructGPT \cite{instructGPT} and ChatGPT \cite{ChatGPT-OpenAI}, we design the \textbf{task-related instruction} describing the relation extraction task and add it to the prompt, which is named \textbf{\textsc{instruct prompt}}.
Meanwhile, according to previous few-shot RE works \cite{typmarker}, entity types (schemas) are helpful; therefore, we also explore the effectiveness of schemas in prompts.

\subsection{Data Generation with LLMs}
\label{sec:DA}
To complement the scarcity of labeled data, we introduce another strategy: data generation via LLMs.
Specifically, we utilize specific prompts with descriptions of data forms to guide LLMs to generate more in-domain labeled data autonomously, which is subsequently employed to fine-tune a relatively small language model with existing few-shot labeled training data.
We design the prompt to tell the essential components ($\boldsymbol{x}, \boldsymbol{h}, \boldsymbol{t}, \boldsymbol{t_h}, \boldsymbol{t_t} \ \text{and} \ \boldsymbol{y}$) of one RE training instance and show few-shot instances as demonstrations to teach LLMs to comprehend features of labeled RE data.
Note that schemas, such as types of relations and entities, are significant structural information in RE data.
Therefore, we propose \textbf{schema-constrained data generation} by adding entity types as schema guidance to the prompt (in Figure \ref{fig:intro}) to boost performance.
Then, the prompt is utilized to guide LLMs to create augmented relation extraction data that are converted into the expected format for future usage.

\section{Experimental Setups}
\subsection{Methods and Datasets}
\textbf{GPT-3.5} is utilized via OpenAI API\footnote{\url{https://platform.openai.com/docs/models/gpt-3-5}} as the large language model in our experiments.
We implement experiments on four relation extraction datasets, including TACRED \cite{tacred}, TACREV \cite{tacrev}, RE-TACRED \cite{retacred} and SciERC \cite{SciERC}.
Compared with the LLM, six baselines methods are conducted via relatively small models (details in Appendix \ref{app:detail}).

\subsection{Few-shot Settings}
\textbf{$K$ instances per relation ($K$-shot) are sampled for training and validation.} 
For all baselines, we use randomly sampled 8-shot and 16-shot datasets for training and validation.
As for in-context learning, because GPT-3.5 has the limitation of maximum request tokens (4097 tokens) and the series of TACRED datasets have more than 40 relations, \textbf{one-shot demonstrations} can only be used, and the one-shot performance is reported in Table \ref{tab:main}.
For the same reason, to generate more labeled data for each relation independently, only three demonstrations for the relation are added to the prompts.

In-context learning is implemented on the four whole test sets. Different demonstrations are randomly sampled from the shuffled training set every time to avoid effects from permutations of demonstrations \cite{lu2021fantastically}.
As for data generation, generated data from GPT-3.5 and original few-shot training data are combined to fine-tune two baselines, TYP Marker \cite{typmarker} and KnowPrompt \cite{knowprompt}.
Using different shots of generated data will lead to different results.
Therefore, we increasingly add generated $k$-shot ($k \in \{8, 16, 32, 48\}$) data to the original 8-shot and 16-shot training data respectively and report the best performance over $k$ in Tabel \ref{tab:main}.
More details are shown in Appendix \ref{app:imple}.

\section{Results and Discussion}
\subsection{Main Findings for Relation Extraction}

\paragraph{In-context learning on LLMs can achieve comparable performance for RE with tuning relatively small PLMs.}
From Table \ref{tab:main}, we notice that ICL with only one-shot demonstrations can obtain competitive performance with full parameter tuning-based prompt learning baselines.
Using LLMs via ICL does not necessitate any parameter updates, which contains the potential value of making models scenario-adaptable, unlike supervised learning requiring parameter optimization.

\paragraph{Data generation with LLMs can boost previous solutions to obtain new state-of-the-art few-shot RE results.}
We find that previous baselines can significantly improve with \textbf{10.7\%} for 16-shot in SciERC and \textbf{6.6\%} for 16-shot in RE-TACRED by simply using generated data from GPT-3.5 in Table \ref{tab:main}.
To be noted, data generation is a simple yet effective approach to elicit the power from the LLM to previous methods, and we demonstrate that using schema-constrained generation with LLMs can benefit all previous approaches with SLMs.

\subsection{Prompts in In-context Learning with LLMs}
\paragraph{Instructions and schemas play an essential role in in-context learning for RE with LLMs.}
From Table \ref{tab:prompt}, we notice that the model with \textsc{instruct prompt} obtains better performance than \textsc{text prompt} in most cases, indicating task-related information indeed helps to unlock more ability of LLMs for RE. 
Aberrant results are shown in TACRED and TACREV because incorrectly labeled demonstrations from the two datasets will violate the correct instruction fed into LLMs, which confuses LLMs and results on worse performance than ICL without the instruction.
Moreover, adding schema information obtains much better performance, exhibiting the importance of pre-defined structural information for relation extraction.

\begin{figure}[t!]
    \centering
    \includegraphics[width =0.87\linewidth]{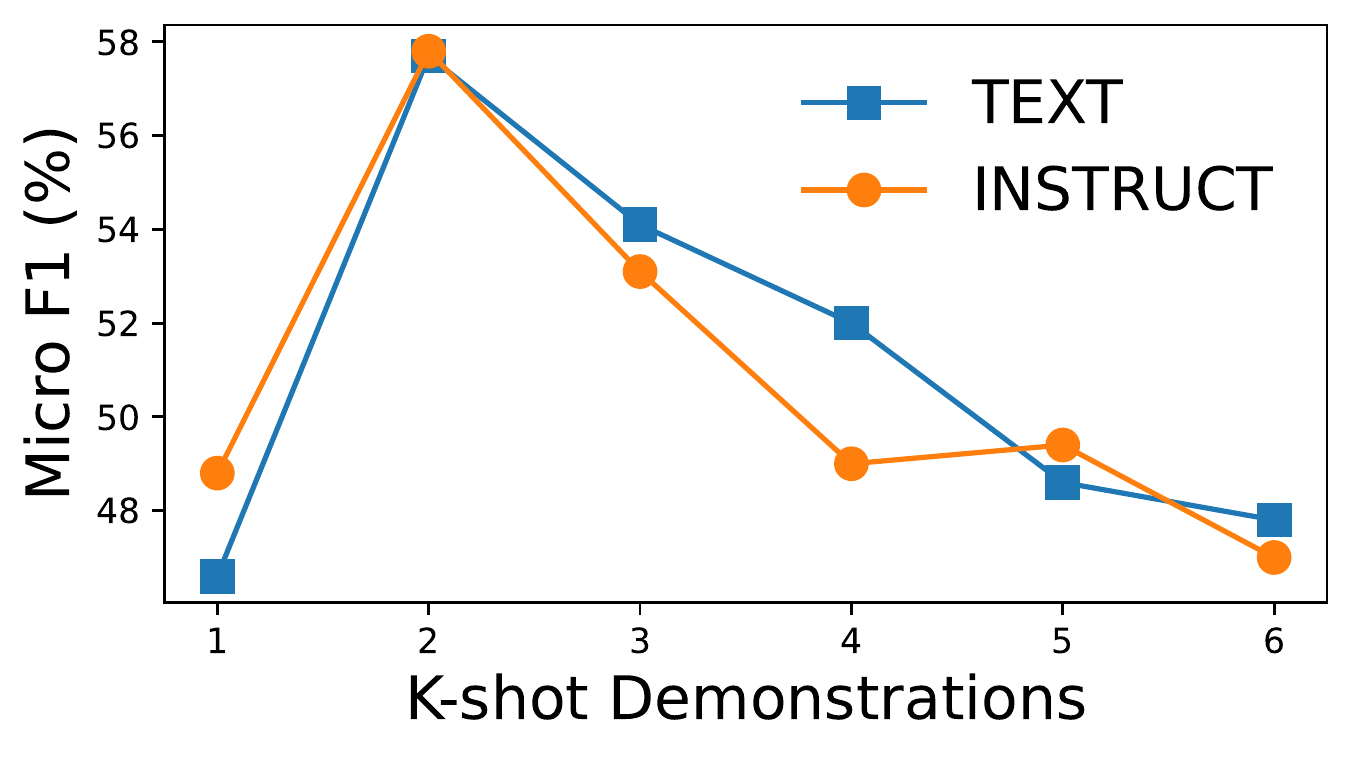}
    \caption{
    Micro F1 (\%) of $k$ in-context demonstrations in SciERC.}
    \label{fig:scidemo}
\end{figure}

\paragraph{More demonstrations, counter-intuitively, may not lead to performance improvement for RE with LLMs.}
We find performance will not improve even drop and the gap between \textsc{instruct prompt} and \textsc{text prompt} becomes relatively smaller as the number of in-context demonstrations increases from Figure \ref{fig:scidemo}.
We argue that there may be two reasons: 1) it is challenging to select representative demonstrations; 2) it is non-trivial for LLMs to understand structure prediction tasks with more large output (relation) space. More case studies for GPT-3.5 can be found in Appendix \ref{app:wrong}.

\subsection{Utility of Generated Data from LLMs}

\paragraph{Combining data generated from LLMs with original training data can yield better RE performance than from traditional data augmentation approaches.}
We compare data generation through the LLM with previous widely used data generation approaches, such as substituting words in training sets with WordNet’s synonyms and contextual word embedding in Figure \ref{fig:different-da} (details in Appendix \ref{app:imple}). Data generation with LLMs can obtain better performance than all others, indicating guiding LLMs to generate data is an effective method to compensate for the lack of labeled data.

\begin{figure}
\centering
\includegraphics[width=0.46\textwidth]{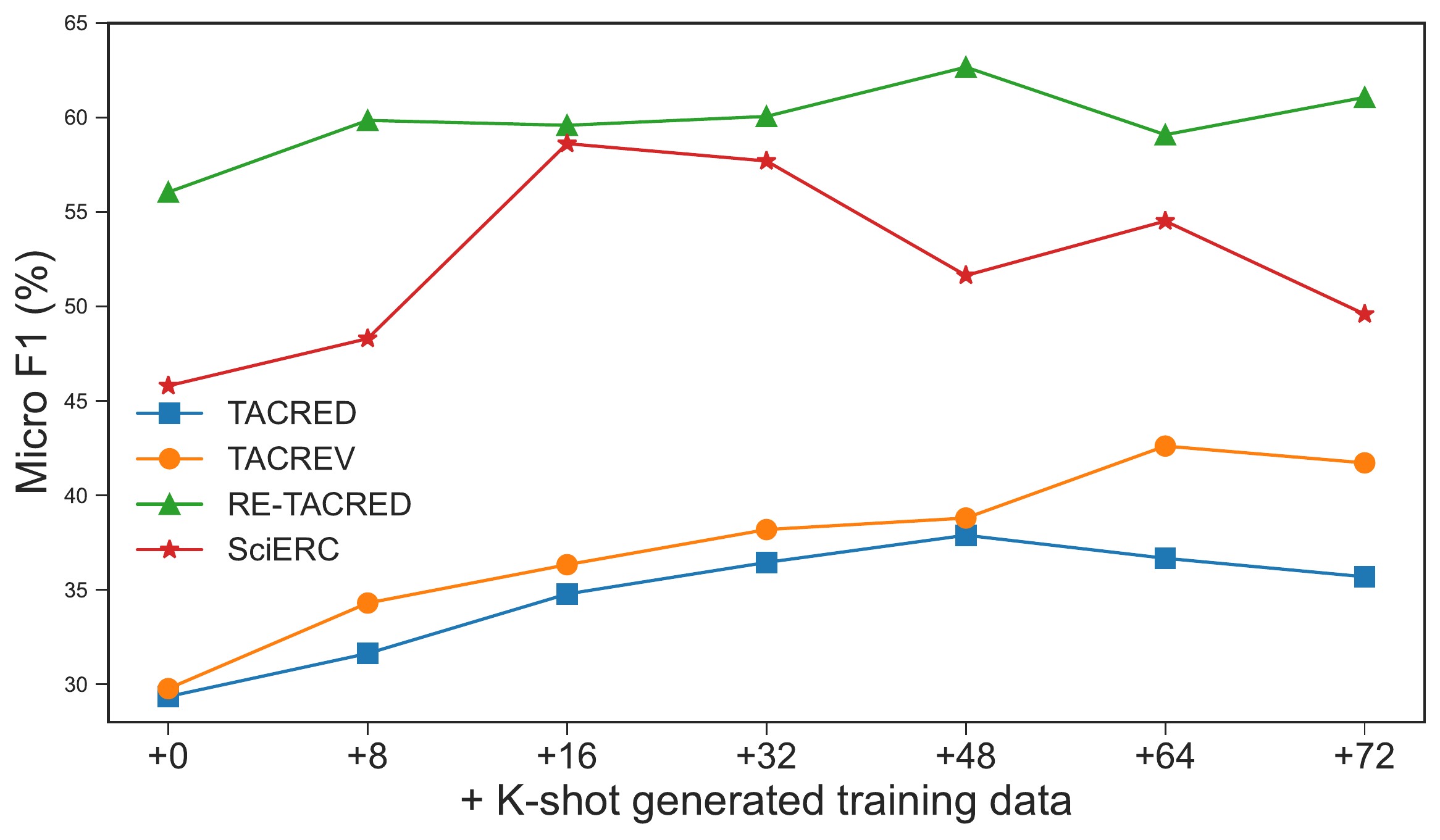}
\caption{Micro F1 (\%) of KnowPrompt with generated training data and original 8-shot data.}
\label{fig:RE}
\end{figure}

\paragraph{Using more and more generated data from LLMs can only boost RE performance to a certain extent, not continuously better.}
From Figure \ref{fig:RE}, we observe that with more generated data, the result climbs up first and then declines, and is always higher than without generated data.
We think low-quality generated data introduces much noise in the training course, according to the analysis on generated data in Appendix \ref{app:error}, and LMs may have an anti-noise capacity \cite{DBLP:journals/corr/abs-2007-08199}.

\begin{figure}
\centering
\includegraphics[width=0.45\textwidth]{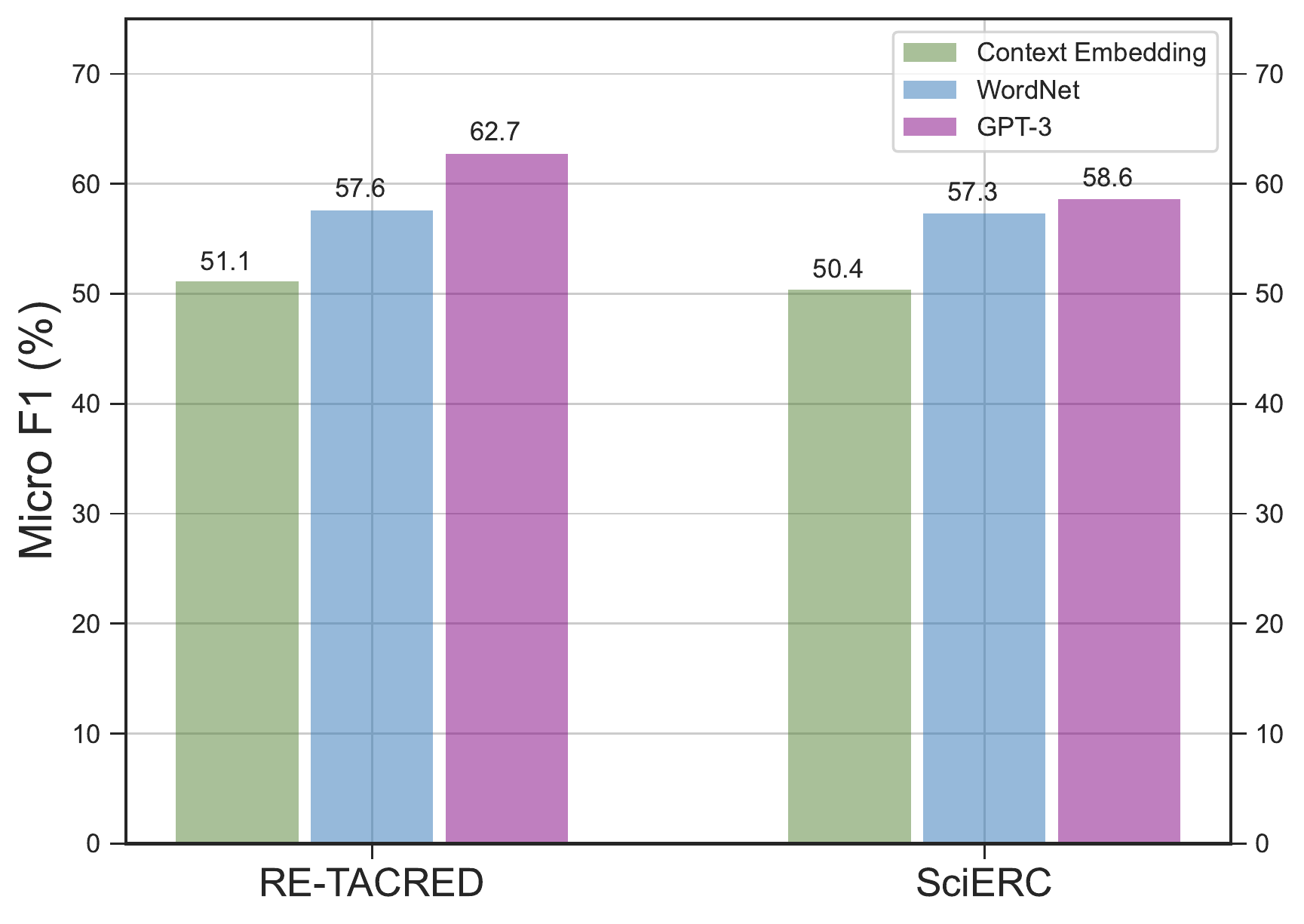}
\caption{Performance of data generation with LLMs and different data augmentation methods. 
\textit{Roberta} and \textit{SciBERT} are used on RE-TACRED and SciERC, respectively, in the context embedding-based DA method.}
\label{fig:different-da}
\end{figure}

\section{Discussion and Conclusion}
In this paper, we take the first step to investigate how to utilize the large language model for few-shot relation extraction.
We observe that task-related information, including instructions or schemas, helps to elicit the capability of LLMs and boost few-shot relation extraction performance.
At this stage, using LLMs to generate data may be a simple yet effective solution to enhance the power of foundation models (relatively small PLMs) for practical applications.
We hope this work can deliver the benefits of using LLMs for the NLP community.
Note that LLMs can make predictions only based on contexts combined with a few training examples as demonstrations.
We argue that it has the potential to design sophisticated human-readable prompts for scenario-adaptable (e.g., low-shot and any domains) relation extraction. 

\section*{Acknowledgment}

We would like to express gratitude to the anonymous reviewers for their kind comments. 
This work was supported by the National Natural Science Foundation of China (No.62206246), Zhejiang Provincial Natural Science Foundation of China (No. LGG22F030011), Ningbo Natural Science Foundation (2021J190), and Yongjiang Talent Introduction Programme (2021A-156-G), CAAI-Huawei MindSpore Open Fund.

\section*{Limitations}

Despite our best efforts, there may still be some limitations remaining in this paper.

\paragraph{LLMs:}
Due to the limited budgets, we can not afford all kinds of LLMs, so we only evaluate GPT-3.5  (\textit{text-davinci-003}).
We will try to investigate relation extraction with more LLMs, such as OPT \cite{DBLP:journals/corr/abs-2205-01068}, GLM-130B \cite{DBLP:journals/corr/abs-2210-02414}, or code language models \cite{DBLP:journals/corr/abs-2304-09048} like Codex.

\paragraph{Other Methods to utilize LLMs:} 
There are several other techniques to leverage LLMs, such as black-box optimization \cite{DBLP:conf/icml/SunSQHQ22} and feature-based learning \cite{DBLP:conf/icml/LangAKS22}; however, we find that most of those approaches cannot directly be applied to relation extraction due to the large label space and complex schema structures.
We leave these for future work to leverage other methods with LLMs for relation extraction. 

\paragraph{Datasets:}
 We only evaluate four relation extraction datasets and will try to investigate relation extraction performance with LLMs on more diverse datasets across different domains and languages. 

\bibliography{anthology,custom}
\bibliographystyle{acl_natbib}

\appendix
\section{Experimental Details}
\label{app:detail}
\subsection{Datasets}
\label{app:data}
TACRED\footnote{\url{https://nlp.stanford.edu/projects/tacred/}} is a widely used RE dataset. It has 42 relation labels, including \textit{no\_relation}, meaning no relation is found.
TACREV\footnote{\url{https://github.com/DFKI-NLP/tacrev}} includes the same training set and relabeled development and test sets from TACRED.
RE-TACRED\footnote{\url{https://github.com/gstoica27/Re-TACRED}} is a re-annotated version of TACRED with 40 relations.
SciERC\footnote{\url{http://nlp.cs.washington.edu/sciIE/}} has seven relation categories and is constructed in the scientific domain.
All datasets are derived from their official webs without modification, including contents and train/test/dev splits.
\subsection{Baselines}
\label{app:baseline}
We compare LLMs with recent baseline methods using relatively small models.
1) Normal fine-tuning methods: \textbf{SpanBERT} \cite{spanbert}, a span-based PLM; \textbf{LUKE} \cite{luke}, pre-trained contextualized representations of words and entities based on the bidirectional transformer; \textbf{GDPNet}, a gaussian dynamic time warping pooling net able to select important words for relation prediction; \textbf{TYP Marker} \cite{typmarker}, fine-tuning with entity typed markers.
2) Generative method: \textbf{TANL} \cite{tanl}, framing a structured prediction language task as a translation task between augmented natural languages.
3) Prompt-tuning methods: \textbf{KnowPrompt}, knowledge-aware continuous prompt-based tuning with synergistic optimization.

\begin{table*}[htbp]
  \centering
  \scalebox{0.9}{
    \begin{tabular}{lcccccccc}
    \toprule
    \multicolumn{1}{r}{\multirow{2}[2]{*}{8-shot Dataset}} & \multicolumn{2}{c}{\textbf{TACRED}} & \multicolumn{2}{c}{\textbf{TACREV}} & \multicolumn{2}{c}{\textbf{RE-TACRED}} & \multicolumn{2}{c}{\textbf{SciERC}} \\
      & generated & gold & generated & gold & generated & gold & generated & gold \\
    \midrule
    add 0-shot & 29.35 & 29.35 & 29.77 & 29.77 & 56.05 & 56.05 & 45.80 & 45.80 \\
    add 8-shot & 31.63 & 30.73 & 34.30 & 33.16 & 59.85 & 60.92 & 48.30 & 57.08 \\
    add 16-shot & 34.78 & 31.88 & 36.33 & 33.49 & 59.59 & 61.30 & \textbf{58.62} & 65.15 \\
    add 32-shot & 36.45 & 33.35 & 38.19 & 33.98 & 60.06 & 64.65 & 57.70 & 72.11 \\
    add 48-shot & \textbf{37.89} & 33.97 & 38.80 & 35.06 & \textbf{62.67} & 65.56 & 51.64 & 74.29 \\
    add 64-shot & 36.67 & 34.36 & \textbf{42.61} & 35.57 & 61.07 & 67.28 & 54.52 & 75.36 \\
    add 72-shot & 35.69 & \textbf{34.58} & 41.72 & \textbf{35.96} & 59.09 & \textbf{67.43} & 49.59 & \textbf{75.87} \\
    \bottomrule
    \end{tabular}}
    \caption{Micro F1 (\%) of \textit{KnowPrompt} after adding labeled data generated by GPT-3.5 or gold labeled data to 8-shot datasets.}
  \label{tab:generate_vs_gold}
\end{table*}

\subsection{Implementation Details}
\label{app:imple}
Generated data with existing training data is then evaluated on KnowPrompt.
Data augmentation methods with Word-Net’s synonyms and contextual word embedding are achieved by \textit{nlpaug}\footnote{\url{https://github.com/makcedward/nlpaug}}.
The parameter \emph{temperature} in OpenAI API is set to 0 for precision in ICL and 1 for generating diverse RE data.
One NVIDIA GeForce RTX 3090 GPU with 24GB memory is employed to run all experiments.
We rerun the official code of baselines with their original settings except on the SciERC dataset.
Due to the vertical domain of SciERC, \textit{SciBERT} \cite{scibert} is used in TYP Marker and KnowPrompt for fairness.
And for another three datasets, \textit{RoBERTa-large} is utilized in \textit{TYP Marker} and \textit{KnowPrompt}.

\section{Case Analysis}
\label{app:caseana}
\subsection{Wrong Cases from ICL}
\label{app:wrong}
\begin{table*}[htbp]
  \centering
  \scalebox{0.74}{
    \begin{tabular}{cp{22.4em}ll}
    \toprule
    Dataset & \multicolumn{1}{c}{Case} & \multicolumn{1}{c}{Gold Relation} & \multicolumn{1}{c}{In-context Learning} \\
    \midrule
    \multirow{2}[4]{*}{TACRED} & Context: And strangely enough , Cain's short , three-year tenure at the NRA is evidently the only period in his decades-long career during which he 's alleged to have been a sexual predator.\newline{}Head Type: ORGANIZATION. Head Entity: NRA.\newline{}Tail Type: PERSON. Tail Type: Cain & org:top\_members/employees & per:employee\_of \\
\cmidrule{2-4}      & Context: "I learn from students and I challenge them," says Heloise, 58, who took over the family hints business when her mother, also named Heloise, died in 1977.\newline{}Head Type: PERSON. Head Entity: Heloise.\newline{}Tail Type: PERSON. Tail Entity: Heloise. & per:parents & per:alternate\_names \\
    \midrule
    \multirow{2}[4]{*}{TACREV} & Context: Anna Mae Pictou Aquash, a Mi ` kmaq Indian from Canada, was brutally murdered in 1975.\newline{}Head Type: PERSON. Head Entity: Anna Mae Pictou Aquash.\newline{}Tail Type: COUNTRY. Tail Entity: Canada. & per:country\_of\_birth & per:countries\_of\_residence \\
\cmidrule{2-4}      & Context: Messina Denaro has been trying to impose his power in Palermo, the Sicilian capital, and become the new head of the Sicilian Mafia, weakened by the arrest of Provenzano in April 2006.\newline{}Head Type: PERSON. Head Entity: his.\newline{}Tail Type: CITY. Tail Entity: Palermo. & no\_relation & per:cities\_of\_residence \\
    \midrule
    \multirow{2}[4]{*}{RE-TACRED} & Context: They say Vladimir Ladyzhenskiy died late Saturday during the contest in southern Finland, while his Finnish rival Timo Kaukonen was rushed to a hospital.\newline{}Head Type: PERSON. Head Entity: Vladimir Ladyzhenskiy.\newline{}Tail Type: PERSON. Tail Entity: his. & per:identity & per:date\_of\_death \\
\cmidrule{2-4}      & President of the Central American Parliament (Parlacen) Jacinto Suarez said on Monday that the presidents of the Central American countries did not support Panama 's request of withdrawal from the Parlacen.\newline{}Head Type: ORGANIZATION. Head Entity: Central American Parliament.\newline{}Tail Type: PERSON. Tail Entity: Jacinto Suarez. & org:top\_members/employees & per:title \\
    \midrule
    \multirow{2}[4]{*}{SciERC} & Context: We evaluate across two corpora (conversational telephone speech and broadcast news speech) on both human transcriptions and speech recognition output.\newline{}Head Type: OtherScientificTerm. Head Entity: transcriptions.\newline{}Tail Type: OtherScientific Term. Tail Entity: output. & CONJUNCTION & COMPARE \\
\cmidrule{2-4}      & Context: We validate this new method on nine standard person re-identification datasets including two large scale Market-1501 and CUHK03 datasets and show that we improve upon the current state-of-the-art methods on all of them.\newline{}Head Type: Material. Head Entity: CUHK03 datasets.\newline{}Tail Type: Material. Tail Entity: datasets. & HYPONYM-OF & PART-OF \\
    \bottomrule
    \end{tabular}}
    \caption{Wrong cases predicted by GPT-3.5. The gold relation categories are listed in the third column and the results predicted by in-context learning are in the fourth column.}
  \label{tab:wrong}
\end{table*}

From Table \ref{tab:wrong}, we notice that some RE instances are challenging for LLMs, and there are several limitations with LLMs: 
1) LLMs are not good at clearly distinguishing the order between head and tail entities.
2) The same mention of head and tail entities will confuse LLMs.
4) If the distance between head and tail entities in the context is long, it is difficult for LLMs to decide the relation correctly.
5) Semantically-similar relation label words and entity mentions will puzzle LLMs because their embeddings are similar.
6) LLMs cannot afford very long instances since there is a large label space for relation extraction. 
7) LLMs may mostly fail to extract those ambitious or wrongly labeled relations; those are also challenging for humans. 
More high-quality demonstrations may help mitigate these issues. 
And we think it is necessary to develop step-by-step (Chat-style) approaches with LLMs to extract limited relations in one stage.

\subsection{Generated Data from LLMs}
\label{app:error}
There are some cases for generated data from GPT-3.5 in Table \ref{tab:gendata}.
Through human checks on 100 generated samples per dataset, about 78\% generated data are corrected labeled and of a high quality (85\% for TACRED, 82.5\% for TACREV, 72\% for RE-TACRED, 75\% for SciERC).
Meanwhile, we add generated data and original gold training data respectively to 8-shot datasets and fine-tune \textit{KnowPrompt}, we evaluate the quality of generated data as shown in Table \ref{tab:generate_vs_gold}.
We observe that labeled data generated by GPT-3.5 are mostly correct. As for TACRED and TACREV, generated data achieve more improvements than gold labeled data. Since there are many incorrect labeled data in TACRED and TACREV \cite{tacred, tacrev}, we think better performance results from GPT-3.5's help.
However, we also find that Some generated data from GPT-3.5 are of less quality than gold data.
As for RE-TACRED and SciERC, using more gold data perform better than generated data. Through human checks, some generated samples are too short and concatenated by some semantically irrelevant sentences. Meanwhile, big performance's difference on SciERC shows GPT-3.5 is not good at vertical domains such as science.

\begin{table*}[htbp]
  \centering
    \scalebox{0.74}{
    \begin{tabular}{cp{30.5em}l}
    \toprule
    Dataset & \multicolumn{1}{c}{Case} & \multicolumn{1}{c}{Corrective Data} \\
    \midrule
    \multirow{2}[4]{*}{TACRED} & Context: The American Cancer Society is headquartered in Atlanta and was founded in 1913 by 15 trained laywomen.\newline{}Head Type: ORGANIZATION. Head Entity: American Cancer Society.\newline{}\textbf{Tail Type: ORGANIZATION}. Tail Entity: 15 trained laywomen.\newline{}Relation: org:founded\_by. & Tail Type: PERSON \\
\cmidrule{2-3}      & Context: Mary Brown, CEO of Brown Corp and renowned businesswoman, is a regular speaker at industry conferences and events.\newline{}Head Type: PERSON. Head Entity: Mary Brown.\newline{}\textbf{Tail Type: PERSON}. Tail Entity: CEO.\newline{}Relation: per:title. & Tail Type: TITLE \\
    \midrule
    \multirow{2}[4]{*}{TACREV} & Context: Gustav Mahler was born in Kalischt, Bohemia on July 7th, 1860.\newline{}Head Type: PERSON. Head Entity: Gustav Mahler.\newline{}\textbf{Tail Type: PERSON}. Tail Entity: 1860.\newline{}\textbf{Relation: per:country\_of\_birth}. & \multicolumn{1}{p{12.565em}}{Tail Type: DATE\newline{}Relation: per:date\_of\_birth} \\
\cmidrule{2-3}      & Context: MTN Nigeria, a subsidiary of South African-based MTN Group, has begun to list its shares on the Nigerian Stock Exchange.\newline{}Head Type: ORGANIZATION. Head Entity: MTN Group.\newline{}Tail Type: ORGANIZATION. Tail Entity: MTN Nigeria.\newline{}Relation: org:subsidiaries. & - \\
    \midrule
    \multirow{2}[4]{*}{RE-TACRED} & Context: Pope John Paul II was a hugely popular Catholic leader who was based in the Vatican City for most of his papacy.\newline{}Head Type: PERSON. Head Entity: Pope John Paul II.\newline{}\textbf{Tail Type: PERSON}. Tail Entity: Vatican City.\newline{}\textbf{Relation: per:countries\_of\_residence}. & \multicolumn{1}{p{12.565em}}{Tail Type: CITY\newline{}Reltaion: per:cities\_of\_residence} \\
\cmidrule{2-3}      & Context: French drug manufacturer Sanofi-Aventis dissolved its Chinese subsidiary Guangzhou Pharma following a bribery scandal.\newline{}Head Type: ORGANIZATION. Head Entity: Sanofi-Aventis.\newline{}Tail Type: ORGANIZATION. Tail Entity: Guangzhou Pharma.\newline{}Relation: org:dissolved. & - \\
    \midrule
    \multirow{2}[4]{*}{SciERC} & Context: The comparison between the two approaches indicates that the neural method produces far better results than the rule-based system.\newline{}Head Type: Method. Head Entity: neural method.\newline{}Tail Type: Method. Tail Entity: rule-based system.\newline{}Relation: COMPARE. & - \\
\cmidrule{2-3}      & Context: The combination of chromatography and mass spectrometry has enabled scientists to achieve unparalleled levels of proteome analysis.\newline{}Head Type: Method. Head Entity: mass spectrometry.\newline{}Tail Type: Method. Tail Entity: chromatography.\newline{}\textbf{Relation: FEATURE-OF}. & Relation: CONJUNCTION \\
    \bottomrule
    \end{tabular}}
    \caption{Generated data from LLMs. Errors are bold in the second column and corrected in the third column.}
  \label{tab:gendata}
\end{table*}

\end{document}